\documentclass{article}
\usepackage{spconf,amsmath,epsfig}
\usepackage{multirow}
\usepackage{amsfonts}
\title{VQA-Aid: Visual Question Answering for Post-Disaster Damage Assessment and Analysis}
\name{Argho Sarkar, Maryam Rahnemoonfar\thanks{This work is partially supported by Microsoft.}}
\address{Bina Lab, University of Maryland, Baltimore County\\ Maryland, USA}
\begin{document}
%\ninept
%
\maketitle
\begin{abstract}

Visual Question Answering system integrated with Unmanned Aerial Vehicle (UAV) has a lot of potentials to advance the post-disaster damage assessment purpose. Providing assistance to affected areas is highly dependent on real-time data assessment and analysis. Scope of the Visual Question Answering is to understand the scene and provide query related answer which certainly faster the recovery process after any disaster. In this work, we address the importance of \textit{visual question answering (VQA)} task for post-disaster damage assessment by presenting our recently developed VQA dataset called \textit{HurMic-VQA} collected during hurricane Michael, and comparing the performances of baseline VQA models.

\end{abstract}
\section{Introduction}
\label{sec:intro}
\begin{figure}[h]
\centering
        \includegraphics[scale=.55]{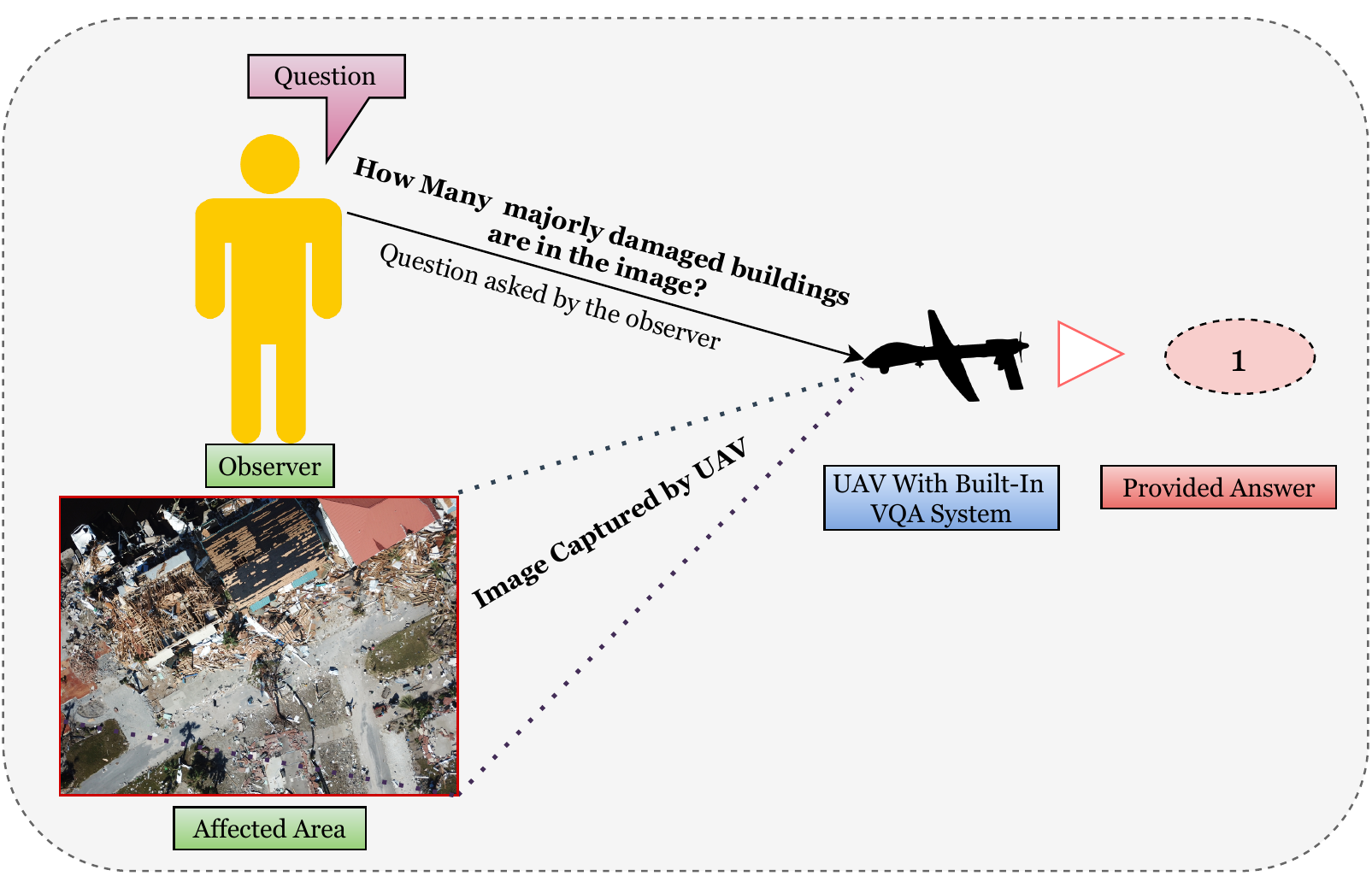}
   
    %    \source{Source:https://www.beckmancoulter.com/en/products/hematology/early-sepsis-detection}
    \caption{\textit{VQA-Aid}: At first, a UAV with built-in VQA system captures the images from the affected region and an observer asks a question relevant to the scenario. Finally, after the analysis, the device produces responses.}
    \ \label{Fig:system}
\end{figure}

\begin{figure*}[h]
\centering
        \includegraphics[scale=.65]{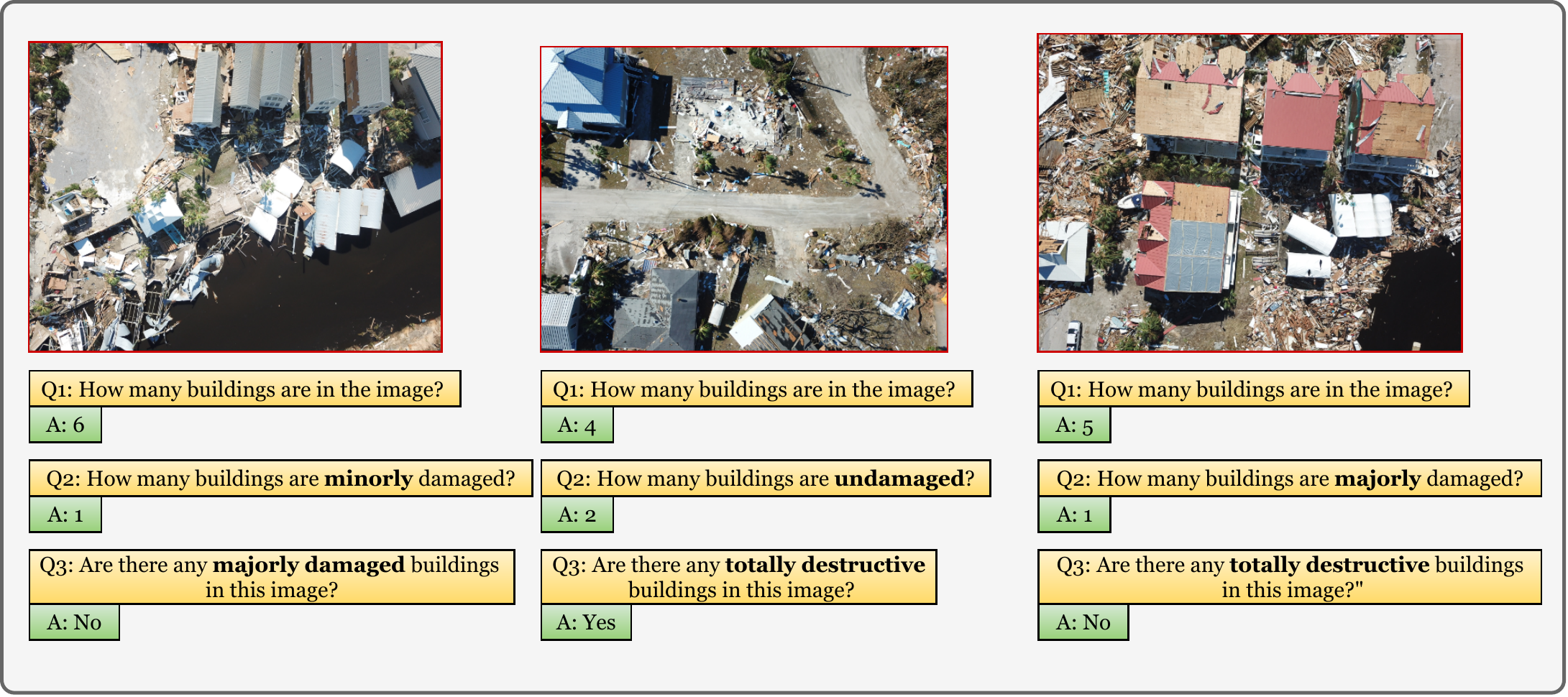}
 
    %    \source{Source:https://www.beckmancoulter.com/en/products/hematology/early-sepsis-detection}
    \caption{For all the images, \textit{Q1} represents the \textit{simple counting} question.\textit{ Q2} and \textit{Q3} are the reflection  of \textit{complex counting} and \textit{yes/no} type of questions. We aim to count the object of a particular attribute in \textit{complex counting} questions (e.g. number of \textit{majorly/minorly} damaged buildings instead of total number of building).   }
      \label{Fig:data_rep}
\end{figure*}

Visual Question Answering (VQA) is a complicated multimodal research problem in which the aim is to address an image-specified question. Thus, to find the right answer, VQA systems need to model the question and image (visual content). Visual Question Answering is regarded as a cognitive activity that separates it from other perceptual activities, such as the classification of images. For providing answers based on questions in natural language, a VQA model needs to identify the relevant objects from the images, recognize the attributes and find out the interactive relationships among several objects. This high-level scene understanding has the potential to advance the decision support systems for post-disaster damage assessment.  Answers from the questions such as  ``What is the condition of the road?'', ``How many buildings are damaged?'' provides vital information that assists and faster the recovery process which could save many lives. Additionally, the management and the distribution of limited resources can be allocated optimally with the information from the VQA system. However, the success of any VQA model depends on the task-specific data. As the collection of the data is laborious as well as risky due to difficulties to enter the affected areas because of many adverse conditions such as damaged roads, flooded areas, etc., an automated system such as UAV integrated with the VQA module, trained on disaster specific dataset, can be implemented for damage assessment purpose. Understanding the scarcity of VQA datasets for post-disaster damage assessment, we develop a VQA dataset namely \textit{HurMic-VQA} collected after the \textit{Hurricane Michael}. 
Figure \ref{Fig:system} represents the \textit{VQA-Aid} framework in which we showed how the VQA task can be introduced as an assistant tool for disaster assessment that enables us to make the right decision at any time.

\par Although several datasets are provided for post-disaster damage assessment purposes. Most of those datasets \cite{bischke2017multimedia,benjamin2018multimedia,gupta2019xbd} contain satellite images and images collected from social media. However, in \cite{rahnemoonfar2020floodnet} authors provide high resolution UAV images.  Satellite images are usually captured from high altitudes therefore they have low resolution. Our \textit{HurMic-VQA} dataset contains high resolution UAV images.  In most cases, tasks related to the available datasets for natural disaster are limited to classification \cite{gupta2019creating, kyrkou2019deep} and semantic segmentation \cite{gupta2019creating, rudner2019multi3net,9377916}.  Visual Question Answering practice has not been considered much for the post-disaster damage assessment under climate change issue.  To the best of our knowledge, this is the first work addressing VQA tasks based on UAV imagery in climate issues. Substantial research efforts have been made on the development of VQA algorithms \cite{zhou2015simple, antol2015vqa, chen2015abc, shih2016look, yang2016stacked, fukui2016multimodal, kim2016hadamard, yu2017multi, lu2016hierarchical} in the computer vision and natural language processing communities on many datasets \cite{10.1007/978-3-642-33715-4_54,antol2015vqa}.  In these methods, different approaches for the fined-grained fusion between semantic image and question features have been proposed \cite{zhou2015simple,yu2017multi,yang2016stacked,lu2016hierarchical}. However, the implementation of  VQA algorithms for  UAV imagery is complex compare to the other datasets. The representation of UAV images is vertical which is opposite from the everyday images. Differentiating among several objects from a high altitude makes it difficult even for a human. The scenario of the affected areas after a disaster makes it more complicated as there exist many noises such as debris from structural damage compare to the pre-disaster condition. No benchmark results of the well-established  VQA algorithms have been provided regarding post-disaster damage assessment based on UAV imagery. To address this issue, we compare the baseline VQA models, in this work, on our dataset. 
Our work is unique for two reasons. Firstly, we introduce the VQA dataset for post-disaster damage assessment based on UAV imageries, and finally, we conduct a comprehensive study of the performances of baseline VQA algorithms on our dataset.

\section{Dataset Description}
\label{ssec:subhead}
% \begin{table}[h]
% \begin{tabular}{|c|c|c|c|c|}
% \hline
% \textbf{Data Set} &
%   \textbf{Overall Accuracy} &
%   \textbf{\begin{tabular}[c]{@{}c@{}}Accuracy for \\ “Yes/No”\end{tabular}} &
%   \textbf{\begin{tabular}[c]{@{}c@{}}Accuracy for \\ “Count”\end{tabular}} &
%   \textbf{\begin{tabular}[c]{@{}c@{}}Accuracy for \\ “Others”\end{tabular}} \\ \hline
% Training   & .682 & .69 & .60 & .73 \\ \hline
% Validation & .677 & .70 & .59 & .71 \\ \hline
% Testing    & .681 & .70 & .60 & .72 \\ \hline
% \end{tabular}
% \end{table}

% \begin{table}[h]
% \begin{tabular}{|c|c|c|c|c|}
% \hline
% \textbf{Data Set} &
%   \textbf{Overall Accuracy} &
%   \textbf{\begin{tabular}[c]{@{}c@{}}Accuracy for \\ “Yes/No”\end{tabular}} &
%   \textbf{\begin{tabular}[c]{@{}c@{}}Accuracy for \\ “Count”\end{tabular}} &
%   \textbf{\begin{tabular}[c]{@{}c@{}}Accuracy for \\ “Others”\end{tabular}} \\ \hline
% Training   & .66 & .69 & .60 & .68 \\ \hline
% Validation & .66 & .70 & .59 & .69 \\ \hline
% Testing    & .67 & .70 & .60 & .69 \\ \hline
% \end{tabular}
% \end{table}
\begin{table*}[htbp]
\centering
\caption{Accuracy Results from Baseline VQA Models}
\vspace{.2cm}
\scalebox{.7}{
\begin{tabular}{|c|c|c|c|c|c|c|}
\hline
Mode of Feature Combination &
  Loss Function &
  Data Type &
  Overall Accuracy &
  \begin{tabular}[c]{@{}c@{}}Accuracy for \\ ``Simple Counting"\end{tabular} &
  \begin{tabular}[c]{@{}c@{}}Accuracy for\\ ``Complex Counting"\end{tabular} &
  \begin{tabular}[c]{@{}c@{}}Accuracy for \\ ``Yes/No"\end{tabular} \\ \hline
\multirow{6}{*}{Concatenation \cite{zhou2015simple}}             & \multirow{3}{*}{Cross Entropy} & Training   & 0.59 & 0.6  & 0.56 & 0.65 \\ \cline{3-7} 
                                           &                                & Validation & 0.57 & 0.58 & 0.54 & 0.60 \\ \cline{3-7} 
                                           &                                & Testing    & 0.55 & 0.56 & 0.53 & 0.59 \\ \cline{2-7} 
                                           & \multirow{3}{*}{KL Divergence} & Training   & 0.6  & 0.6  & 0.56 & 0.67 \\ \cline{3-7} 
                                           &                                & Validation & 0.58 & 0.6  & 0.54 & 0.61 \\ \cline{3-7} 
                                           &                                & Testing    & 0.55 & 0.56 & 0.53 & 0.59 \\ \hline
\multirow{6}{*}{Point-wise Multiplication \cite{antol2015vqa}} & \multirow{3}{*}{Cross Entropy} & Training   & 0.59 & 0.6  & 0.55 & 0.64  \\ \cline{3-7} 
                                           &                                & Validation & 0.58 & 0.61 & 0.54 & 0.61 \\ \cline{3-7} 
                                           &                                & Testing    & 0.57 & 0.56 & 0.53 & 0.65 \\ \cline{2-7} 
                                           & \multirow{3}{*}{KL Divergence} & Training   & 0.6  & 0.6  & 0.56 & 0.67 \\ \cline{3-7} 
                                           &                                & Validation & 0.59 & 0.6  & 0.54 & 0.66 \\ \cline{3-7} 
                                           &                                & Testing    & 0.58 & 0.56 & 0.53 & 0.68 \\ \hline
\multirow{6}{*}{MFB Module \cite{yu2017multi}}                & \multirow{3}{*}{Cross Entropy} & Training   & 0.59 & 0.6  & 0.56 & 0.64 \\ \cline{3-7} 
                                           &                                & Validation & 0.58 & 0.6  & 0.54 & 0.63 \\ \cline{3-7} 
                                           &                                & Testing    & 0.56 & 0.56 & 0.53 & 0.63 \\ \cline{2-7} 
                                           & \multirow{3}{*}{KL Divergence} & Training   & 0.59 & 0.6  & 0.56 & 0.63 \\ \cline{3-7} 
                                           &                                & Validation & 0.57 & 0.6  & 0.54 & 0.60 \\ \cline{3-7} 
                                           &                                & Testing    & 0.57 & 0.56 & 0.53 & 0.64 \\ \hline
\end{tabular}}
\label{res}
\end{table*}
 
% Compared with point-wise multiplication, results are not improved by considering the MFB module as a means of feature combination.

\subsection{Data Collection Process}
The dataset is collected with a small UAV platform, DJI Mavic Pro quadcopters, after \textit{Hurricane Michael}. The dataset consists of video and imagery taken from several flights at Ford Bend County in Texas and other directly impacted areas. All the images are high in resolution, i.e., $4000\times3000$.  The damage and debris situation after the hurricane is presented in Figure \ref{Fig:data_rep}. Though several objects are present, most of the images include debris and buildings. The buildings include both residential and non-residential structures. Table \ref{tab:obj_attri} shows the object types with different attributes. While generating the questions. these attributes are considered.  In this work, we are interested in investigating the structural damage condition for buildings by asking questions for given images. 
\begin{table}[h]
\caption{Object with associated Attributes}
\vspace{.2cm}
\scalebox{.75}{
\begin{tabular}{|c|c|}
\hline
\textbf{Object} & \textbf{Associated Attribute}                                  \\ \hline
Building        & Total Destructive, Majorly Damaged, Minorly Damaged, No damage \\ \hline
Road            & Covered with Debris, Flooded, Undamaged                        \\ \hline
Water           & Covered with Debris, Flood Water, Clean Water                                     \\ \hline
Pools           & Damaged, Undamaged                                             \\ \hline
\end{tabular}}
\label{tab:obj_attri}
\end{table}

\subsection{Question Type}

%   We therefore include complex counting problems in our dataset for UAV imagery to facilitate the research progress. 
\par Questions are grouped into a three-way category of questions, namely \textit{``Simple Counting"},  \textit{``Complex Counting"}, and \textit{``Yes / No"}. We mainly ask the number of presence of an object in \textit{``Simple Counting"} problem regardless of the associated attribute (e.g. \textit{How many buildings are in the images?}). \textit{Yes / No} type questions concentrate on examining whether an object's particular attribute is present. Finally, \textit{complex counting} type of query is explicitly intended to count the existence of a specific attribute of an object (e.g. \textit{How many  \textbf{majorly damaged} buildings are in the images?}). A total of 3197 images are available and each image is connected to all of the 3 types of questions. Figure \ref{Fig:data_rep} represents these three types of questions.

\section{Method}
\label{sec:typestyle}

The baseline models: \textit{simple baseline} and \textit{Multimodal Factorized Bilinear (MFB) baseline} \cite{yu2017multi} have been considered for this task  and all of these models are configured according to our \textit{HurMic-VQA} dataset.  The main pipeline for the aforementioned baseline VQA models consists of image feature extraction, semantic representation of question, and fine-grained combination of these two features. For image and question feature extraction, respectively, VGGNet (VGG 16) and Two-Layer LSTM are taken into account. Image feature vector \textit{I}, \textit{I} $\in$  $\mathbb{R}^m$ where m represents the dimension of image vector  and semantic question feature \textit{Q}, \textit{Q} $\in$ $\mathbb{R}^n$ where n represents dimension of question vector, are combined in a \textit{simple baseline} method by both concatenation and point-wise multiplication. 1024-D image feature vector (from last pooling layer) and 1024-D question vector (from the last word of Two-Layer LSTM) are considered for our study.

\par For the \textit{MFB baseline} approach, authors in \cite{yu2017multi} proposed the MFB module for a fine-grained combination between image and question feature. The MFB module consists of two phases: expanding and squeezing. Image and question feature vector are multiplied point-wise in the expanding process, followed by a dropout layer. In the squeezing step, sum pooling is considered, followed by power and \textit{L-2} normalization layers.

\par Fully-connected and softmax layers are taken into account after the fine-grained combination of the two features in all approaches to model the answers. Given a question and an image, the models will predict the answer to the question by formulating the problem as a classification task (for a given set of answers).

\section{Experiment and Result}
\label{sssec:subsubhead}
\par After the image and question feature extraction from VGGNet and LSTM layer respectively, three modes of feature combination criteria are considered. 1024 dimensional image and question feature vector are combined by concatenation, point-wise multiplication , or MFB module. By considering both cross-entropy and KL divergence loss, all the models are optimized by stochastic gradient descent ( SGD) with batch size 16. The dataset is split into three sets namely training, validation and testing with 60\%, 20\%, 20\% ratio respectively. In the training phase, models are validated by validation dataset via \textit{early stopping} criterion with patience 30. 
% Concatenation refers simply concatenate the 1024 dimension image and question feature vectors. Element-wise multiplication of image and question feature vector is considered for point-wise multiplication approach. Finally, in MFB approach, MFB Module is taken into account for combination of image and question feature. 

% We can see from Table 2 that the loss of KL-divergence functions well in contrast to the loss of Cross Entropy.
\par Our motivation for developing an efficient UAV imagery-based VQA model for post-disaster damage assessment comes from the performances of baseline models in Table \ref{res}. Accuracy is the performance metric that we consider for the VQA task to compare the baseline models. We consider top-1 accuracy for the comparison purpose. If the ground-truth matches the output (which has the highest probability) from a model, the accuracy for any image is 1, otherwise it is 0. Overall accuracy for all these models lies between .54 and .60 which indicates that models hardly understand the scenario for a given question. \textit{Yes/No} type question has higher accuracy compared to other question types. Furthermore, we highlight the difficulties in dealing with the \textit{complex counting} by comparing the accuracy between \textit{simple counting} and \textit{complex counting} problems. Accuracy for the \textit{complex counting} problem is lower among all the question types. This result highlight the performances of baseline VQA models on our dataset.

\section{Conclusion}
Our aim for this analysis is to study the baseline VQA frameworks by presenting our \textit{HurMic-VQA} dataset for post-disaster damage assessment. This study also upholds the importance of implementing Visual Question Answering task for post-disaster damage assessment. We only consider a subset of our dataset that targets only one type of object for this work. From the baseline results, we can understand the importance of developing an effective VQA algorithm for post-disaster damage assessment. 
\bibliographystyle{IEEEbib}
\bibliography{main}

\end{document}